\documentclass{article}
\usepackage{spconf,amsmath,graphicx}
\usepackage{subcaption}
\usepackage{array}
\usepackage{amsfonts}
\usepackage{amssymb}

\title{Synthetic Data Augmentation using GAN  for Improved Liver Lesion Classification}

\name{Maayan~Frid-Adar$^1$ \quad Eyal~Klang$^2$ \quad Michal~Amitai$^2$ \quad Jacob~Goldberger$^3$ \quad Hayit~Greenspan$^1$}
\address{${}^1$Department of Biomedical Engineering, Tel Aviv University, Tel Aviv, Israel. \\
${}^2$Department of Diagnostic Imaging, The Chaim Sheba Medical Center, Tel-Hashomer, Israel. \\
${}^3$Faculty of Engineering, Bar-Ilan University, Ramat-Gan, Israel.}

\begin{document}
%
\maketitle
\begin{abstract}
In this paper, we present a data augmentation method that generates synthetic medical images using  Generative Adversarial Networks (GANs).
 We propose a training scheme that first uses  classical  data augmentation to enlarge the training set and then further enlarges the data size and its diversity by applying GAN techniques for  synthetic data augmentation.
 Our  method is demonstrated on a limited dataset of computed tomography (CT) images of 182 liver lesions (53 cysts, 64 metastases and 65 hemangiomas).
The classification performance using only classic data augmentation yielded 78.6\% sensitivity and 88.4\% specificity. By adding the synthetic data augmentation the results significantly increased to 85.7\% sensitivity and 92.4\% specificity.

\end{abstract}
\section{Introduction}
\label{sec:intro}

One of the main  challenges in the medical imaging domain is how to cope with the small datasets and limited amount of annotated samples, especially when employing supervised machine learning algorithms that require labeled data and larger training examples.
In medical imaging tasks, annotations are made by radiologists with expert knowledge on the data and task and most annotations of medical images are time consuming.
Although public medical datasets are available online, and grand challenges have been publicized,
most datasets are still limited in size and only applicable to specific medical problems.
Collecting medical data is a complex and expensive procedure that requires the collaboration of researchers and radiologists \cite{Greenspan2016overviewDeepMedical}.

Researchers attempt to overcome this challenge by using data augmentation schemes, commonly including  simple modifications of dataset images such as translation, rotation, flip and scale. Using such data augmentation to improve the training process of networks has become a standard procedure in computer vision tasks \cite{Alexnet}. However, the diversity that  can be gained from small modifications of the images (such as small translations and small angular rotations) is relatively small. This motivates the use of synthetic data examples; such samples enable the introduction of more variability and can possibly enrich the dataset further,  in order to improve the system training process.

 A promising approach  for training a model that synthesizes images is known as Generative Adversarial Networks (GANs) \cite{Goodfellow2014GAN}. GANs have gained great popularity in the computer vision community and different variations of GANs were recently proposed for generating high quality realistic natural images \cite{Radford2015DCGAN,Odena2016ACGAN}.
Recently, several medical imaging applications have applied the GAN framework
\cite{costa2017towards,Nie2017,Schlegl2017ganAnomaly}. Most studies have employed the image-to-image GAN technique to create label-to-segmentation translation, segmentation-to-image translation or medical cross modality translations.
Some studies have been inspired by the GAN method for image inpainting.
In the current study  we investigate the applicability of GAN framework to synthesize high quality medical images for data augmentation.
We focus on improving results in the specific task of liver lesion classification.

The liver is one of three most common sites for metastatic cancer along with the bone and lungs. According to the World Health Organization, in 2012 alone, cancer accounted for 8.2 million deaths worldwide of which 745,000 were caused by liver cancer \cite{Ferlay2015cancer}.
There is a great need and interest in developing automated diagnostic tools based on CT images to assists radiologists in the diagnosis of liver lesions.
Previous studies have presented methods for automatic classification of focal liver lesions in CT images
\cite{Gletsos2003,Chang2017,Diamant2017BOVWMI}.

In the current  work we  suggest an augmentation scheme that is based on combination of standard image perturbation and  synthetic liver lesion generation using GAN  for improved liver lesion classification.
 The contributions of this work are the following: synthesis of high quality focal liver lesions from CT images using generative adversarial networks (GANs), design of a CNN-based solution for the liver lesion classification task
and augmentation of the CNN training set using the generated synthetic data - for improved classification results.

\section{Generating Synthetic liver lesions}
\begin{figure}[!t]
\centering
\includegraphics[width=1.8in]{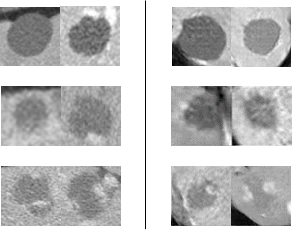}
\caption{Lesion ROI examples of Cysts (top row), Metastases (middle row) and Hemangiomas (bottom row). 
Left side: Real lesions; Right side: Synthetic lesions.}
\label{fig:ROI_real_synth}
\end{figure}

Even a small CNN has thousands of parameters that need to be trained. When using deep networks with multiple layers or dealing  with limited numbers of training images, there is a danger of overfitting.
The standard solution to reduce overfitting is data augmentation that artificially enlarges the dataset \cite{Alexnet}.
Classical augmentation techniques on gray-scale images include mostly affine transformations.
To enrich the training data we apply here an image synthesis technique based on the GAN network. However, to train a GAN we need many examples as well.
The approach we propose here involves several steps: in the first step, standard data augmentation is used to create a larger dataset which  is then used to train a GAN.
The synthetic examples created by the GAN are next used as an additional resource for data augmentation. The combined standard and synthetic augmentation is finally used to train a lesion classifier. 
Examples  of real and synthetic lesions are shown in  Figure \ref{fig:ROI_real_synth}.
We next describe the details of the proposed system.

\subsection{Classic Data Augmentation} \label{sec:data_aug_classic}
 Classic augmentation techniques on gray-scale images include mostly affine transformations such as translation, rotation, scaling, flipping and shearing.
 In order to preserve the liver lesion characteristics we avoided transformations that cause shape deformation (like shearing and elastic deformations). In addition, we kept the ROI centered around the lesion.
Each lesion ROI was first rotated $N_{rot}$ times at random angles $\theta=[0^{\circ},...,180^{\circ}]$. Afterwards, each rotated ROI was flipped $N_{flip}$ times (up-down, left-right), translated $N_{trans}$ times where we sampled random pairs of $[x,y]$ pixel values between $(-p,p)$ related to the lesion diameter (d) by $p=min(4,0.1\times{d})$. Finally the ROI was scaled $N_{scale}$ times from a stochastic range of scales $s=[0.1\times{d},0.4\times{d}]$. The scale was implemented by changing the amount of tissue around the lesion in the ROI. As a result of the augmentation process, the total number of augmentations was $N = N_{rot} \times{(1+N_{flip}+N_{trans}+N_{scale})}$.
Bicubic interpolation was used to resize the ROIs to a uniform  size of $64\times{64}$.

\subsection{GAN Networks for Lesion Synthesis}
\label{sec:DCGAN}

GANs \cite{Goodfellow2014GAN} are a specific framework of a generative model. It aims to implicitly learn the data distribution $p_{data}$ from a set of samples (e.g. images) to further generate new samples drawn from the learned distribution.

\begin{figure*}[!t]
\centering
\includegraphics[width=6in]{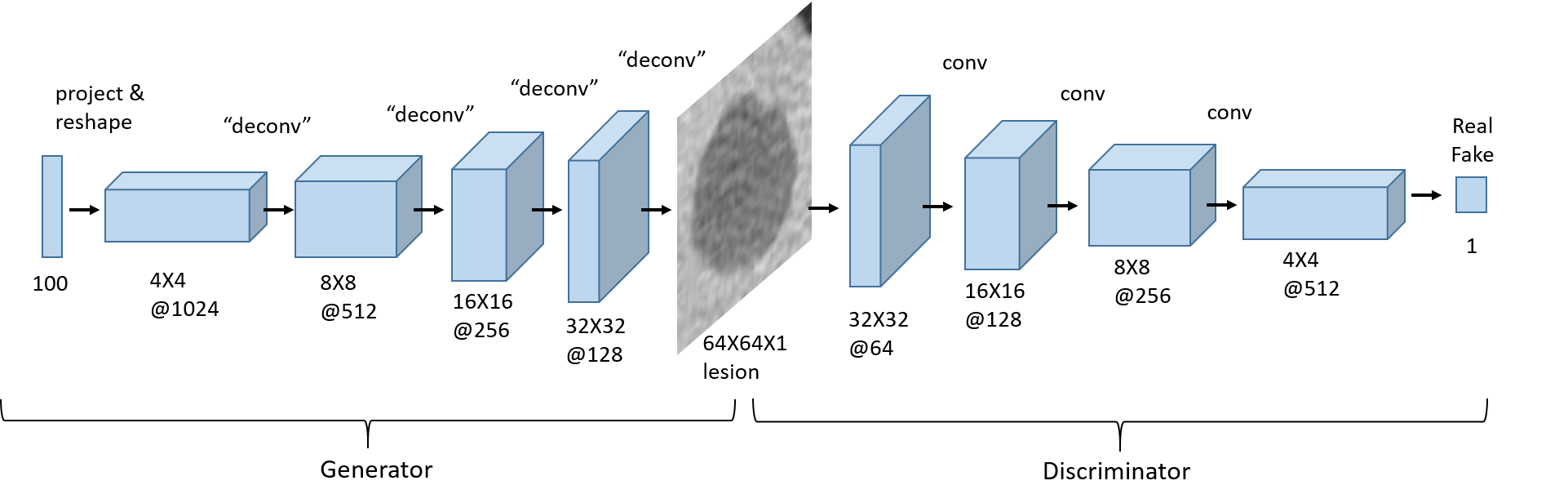}
\caption{Deep Convolutional GAN Architecture (generator+descriminator).}
\label{fig:arch_GAN}
\end{figure*}

We employed the Deep Convolutional GAN (DCGAN) for synthesizing labeled lesions for each lesion class separately:   Cysts, Metastases and Hemangiomas. 
We followed the architecture proposed by Radford et al. \cite{Radford2015DCGAN}.
The model consists of two deep CNNs  that are trained simultaneously, as depicted in  Figure \ref{fig:arch_GAN}.
A sample  $x$ is input to the  discriminator (denoted D), which outputs $D(x)$, its probability of being a real sample.
The generator (denoted G) gets input samples $z$ from a known simple distribution $p_{z}$, and maps $G(z)$ to the image space of distribution $p_{g}$.
During training the generator improves in its ability to synthesize more realistic images while the discriminator improves in its ability to distinguish the real from the synthesized images. Hence the moniker of adversarial training.

     The generator network (Figure \ref{fig:arch_GAN}) takes a vector of 100 random numbers drawn from a uniform distribution as input and outputs a liver lesion image of size $64\times{64}\times{1}$.
     The network architecture \cite{Radford2015DCGAN} consists of a fully connected layer reshaped to size $4\times{4}\times{1024}$ and four \textit{fractionally-strided convolutional} layers to up-sample the image with a $5\times{5}$ kernel size.
 The discriminator network has a typical CNN architecture that takes the input image of size $64\times{64}\times{1}$ (lesion ROI), and outputs a decision - if the lesion is real or fake. In this network, four  convolution layers are used, with a kernel size of $5\times{5}$ and a fully connected layer. \textit{Strided convolutions} are applied to each convolution layer to reduce spatial dimensionality instead of using pooling layers.

\section{Experiments and Results} \label{sec:evaluation_results}

\subsection{Data and Implementation}
\label{sec:data}
The liver lesion data used in this work, was  collected from the Sheba Medical Center.  
Cases  of cysts, metastases and hemangiomas, were acquired from 2009 to 2014 using  a General Electric (GE) Healthcare scanner and a Siemens Medical System scanner, with the following parameters: 120kVp, 140-400mAs and 1.25-5.0mm slice thickness.
Cases were collected with the approval of the institution's Institutional Review Board.

The dataset  was made up of 182 portal-phase 2-D CT scans: 53 cysts, 64 metastases, 65 hemangiomas. An expert radiologist marked the margin of each lesion and determined its corresponding diagnosis which was established by biopsy or a clinical follow-up. This serves as our ground truth.

Liver lesions vary considerably in shape, contrast and size (10-102mm). They also vary in location, where some can be located in interior sections of the liver and some are near its boundary where the surrounding parenchyma tissue of the lesions changes. Finally, lesions  also vary within categories. 
Each type of lesion has its own characteristics but some characteristics may be confusing, in particular for metastasis and hemangioma lesions.
Hemangiomas are benign tumors and metastases are malignant lesions derived from different primary cancers. Thus, the correct identification of a lesion as metastasis or hemangioma is especially important.

We use a  liver lesion classification CNN of the following architecture: three pairs of convolutional layers where each convolutional layer is followed by a max-pooling layer, and two dense fully-connected layers ending with a soft-max layer to determine the network predictions  into the three lesion classes. We use ReLU as activation functions and incorporated a dropout layer with a probability of 0.5 during training.
For training we used a batch size of 64 with a learning rate of 0.001 for 150 epochs.

The input to our classification system are ROIs of $64\times{64}$ cropped from CT scans using the radiologist's annotations.  The ROIs are extracted to capture the lesion and its surrounding tissue relative to its size.
In all experiments and evaluations we used 3-fold cross validation with case separation at the patient level and each fold contained a balanced number of cyst, metastasis and hemangioma lesion ROIs.
For the implementation of the liver lesion classification CNN we used the Keras framework. For the implementation of the GAN architectures we used the TensorFlow framework. All training processes were performed using an NVIDIA GeForce GTX 980 Ti GPU.

\subsection{Evaluation of the Synthetic Data Augmentation} \label{sec:experiment}
We started by examining the effects of using only classic data augmentation for the liver lesion classification task (our baseline). We then synthesized liver lesion ROIs using GAN and examined the classification results after adding the synthesized lesion ROIs to the training set. A detailed description of each step is provided next.

\begin{figure}[h]
\centering
\includegraphics[width=3.6in]{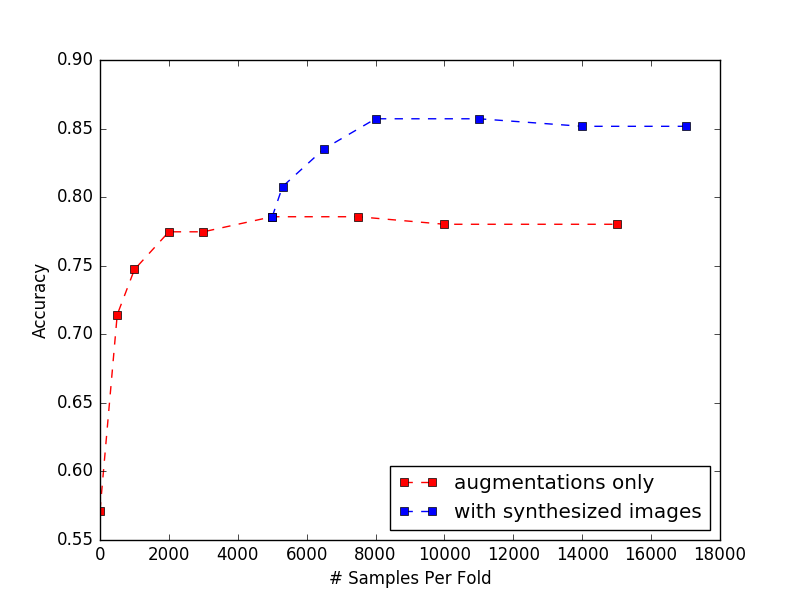}
\caption{Accuracy results for liver lesion classification  with the increase of training set size. The red line shows the effect of adding classic data augmentation and the blue line shows the effect of adding synthetic data augmentation.}
\label{fig:main_results_graph}
\end{figure}

As our baseline, we used classical data augmentation (see section \ref{sec:data_aug_classic}). We refer to this network
as CNN-AUG. We recorded  the classification results for the liver lesion classification CNN for increasing amounts of data augmentation over the original training set.
In order to examine the effect of adding increasing numbers of examples, we formed the data groups  $D_{aug}^{1} \subset D_{aug}^{2} \subset ... \subset D_{aug}^{9}$ such that the first data group was only made up of the original ROIs and each group contains more augmented data. For each original ROI, we produced a large number of augmentations ($N_{rot}=30$, $N_{flip}=3$, $N_{trans}=7$ and $N_{scale}=5$), resulting in $N=480$ augmented images per lesion ROI and overall $\sim 30,000$ examples per folder. Then, we selected the images for the data groups by sampling randomly augmented examples such that for each original lesion we sampled the same augmentation volume.

\begin{table}[!ht]
\renewcommand{\arraystretch}{1.2}
\caption{Confusion Matrix for the Optimal Classical Data
Augmentation Group (CNN-AUG).}
\label{tabel:cm_optimal_augs}
\centering
\begin{tabular}{|c|c|c|c||c|}
\hline
True $\setminus$ Auto & Cyst & Met & Hem & Sensitivity   \\
\hline
Cyst & 52 & 1 & 0 & 98.1\%   \\
\hline
Met & 2 & 44 & 18 & 68.7\%   \\
\hline
Hem & 0 & 18 & 47 & 72.3\%   \\
\hline \hline
Specificity & 98.4\% & 83.9\% & 84.6\% &    \\
\hline
\end{tabular}
\end{table}

\begin{table}[!ht]
\renewcommand{\arraystretch}{1.2}
\caption{Confusion Matrix for the Optimal Synthetic Data Augmentation Group (CNN-AUG-GAN).}
\label{tabel:cm_synth_augs}
\centering
\begin{tabular}{|c|c|c|c||c|}
\hline
True $\setminus$ Auto & Cyst & Met & Hem & Sensitivity   \\
\hline
Cyst & 53 & 0 & 0 & 100\%   \\
\hline
Met & 2 & 52 & 10 & 81.2\%   \\
\hline
Hem & 1 & 13 & 51 & 78.5\%   \\
\hline \hline
Specificity & 97.7\% & 89\% & 91.4\% &    \\
\hline
\end{tabular}
\end{table}

The second step of the experiment consisted of generating synthetic liver lesion ROIs for data augmentation using GAN. We refer to this network
as CNN-AUG-GAN.  Since our dataset was too small for effective training, we incorporated classic augmentation for the training process.
We employed the DCGAN architecture
to train each lesion class separately, using the same 3-fold cross validation process and the same data partition. In all the steps of the learning procedure we maintained a  complete separation between train and test subsets.
After the generator had learned each lesion class data distribution separately, it was able to synthesize new examples by using an input vector of normal distributed samples (``noise").
The same approach that was applied in step one of the experiment when constructing the data groups was also applied in step two: We collected large numbers of synthetic lesions for all three lesion classes, and constructed increased size data groups $\{D_{synth}\}_{j=1}^{6}$ of synthetic examples. To keep the classes balanced, we sampled the same number of synthetic ROIs for each class.

Results of the GAN-based synthetic augmentation experiment are shown in Figure \ref{fig:main_results_graph}.
The baseline results (classical augmentation) are shown in red. We see the total accuracy results for the lesion classification task, for each group of data. When no augmentations were applied, a result of 57\% was achieved; this may be due   to overfitting over the small number of training examples ($\sim 63$ samples per fold). The results improved as the number of training examples increased, up to saturation around 78.6\% where adding more augmented data examples failed to improve the classification results.
We note that the saturation starts with $D_{aug}^{6}=5000$ samples per fold. We define this point as i=optimal where the smallest number of augmented samples were used.
The confusion matrix for the optimal point appears in Table \ref{tabel:cm_optimal_augs}. \\
The blue line in Figure \ref{fig:main_results_graph} shows the total accuracy results for the lesion classification task for the synthetic data augmentation scenario. The classification results significantly improved from 78.6\% with no synthesized lesions to 85.7\% for $D_{aug}^{optimal} + D_{synth}^{3}=5000 + 3000 = 8000$ samples per fold.
The confusion matrix for the best classification results using synthetic data augmentation is presented in Table \ref{tabel:cm_synth_augs}.

\subsection{Expert Assessment of the Synthesized Data}
In order to assess the synthesized lesion data, we challenged two radiologists to classify 
real and fake lesion ROIs into one of three classes: cyst, metastasis or hemangioma.
The goal of the experiment was to check if the radiologists would perform differently on a real lesion vs. a fake one. Similar results would indicate the relevance of the generated data, to the classification task. 

The experts were given, in random order, lesion ROIs from the  original dataset of 182 real lesions  and from  120 additional synthesized lesions.
The expert radiologists' results were compared against the ground truth classification.
It is important to note that in the defined task, we challenged the radiologists to reach a decision based on a single 2-D ROI image. This scenario is not consistent with existing  clinical workflow in which the radiologist makes a decision  based on the entire 3D volume, with support from additional anatomical context, medical history context, and more. 
We are therefore not focusing on the classification results per-se, but rather on the delta in performance between the two presented datasets.

We received the following set of results: 
 Expert 1 classified the real and synthesized lesions correctly in 78\% and 77.5\% of the cases, respectively. Expert 2 classified the real and synthesized lesions correctly in 69.2\% and 69.2\% of the cases, respectively. We observe that for both experts, the classification performances for the real lesions and the synthesized lesions were similar. This suggests that our synthetic generated lesions were meaningful in appearance.

 \section{Conclusion}


  To conclude, in this work we presented a method that uses the generation of synthetic medical images for data augmentation to improve classification performance on a medical problem with limited data. We demonstrated this technique on a liver lesion classification task and achieved a significant improvement of $7\%$ using synthetic augmentation over the classic augmentation.
In the future, we plan to extend our work to additional medical domains that can benefit from synthesis of lesions for improved training - towards increased classification results.

\bibliographystyle{IEEEbib}
\bibliography{strings,refs}

\end{document}